\documentclass{article}
\pdfoutput=1

\usepackage{natbib}
\usepackage{graphicx}
\usepackage{bm}% bold math
\usepackage{algorithm}
\usepackage{algpseudocode}
\usepackage{multicol}

\def\phline{\noalign{\vskip1pt}\hline\noalign{\vskip1pt}}	 % a \hline in a table with a little space
\long\def\sidebyside#1#2{%
\hbox to\textwidth{\vtop{\hsize=.45\textwidth%
\parindent=0pt
\centering
#1\vskip1sp}\hfill\vtop{\hsize=.45\textwidth%
\parindent=0pt
\centering
#2
}}}

\newcommand{\latexOrPdflatex}[2]{\ifx\undefined\pdfoutput%
#1%
\else%
#2%
\fi}

\latexOrPdflatex{
\newcommand{\href}[2]{#2}
}{
\usepackage[colorlinks,pagebackref,citecolor=blue,bookmarks=true,pdfstartview=FitH,pdfstartpage=1,pdftitle={On the numeric stability of the SFA implementation},pdfauthor={Wolfgang Konen},a4paper]{hyperref}
}

\renewcommand{\vec}[1]{\bm{#1}}	% vector boldface
\newcommand{\mat}[1]{\bm{#1}}	% matrix boldface

\newcommand{\is}{x}			% input signal
\newcommand{\tfc}{v}		% transfer function component
\newcommand{\vtfc}{\overline{\vec{\tfc}}}		
\newcommand{\zes}{\overline{\vec{\es}}}		
\newcommand{\es}{z}			% expanded signal
\newcommand{\tf}{g}			% transfer function
\newcommand{\os}{y}			% output signal

\newcommand{\uu}{w}			% time series
\newcommand{\s}{s}			% signal
\newcommand{\p}{\gamma}			% true driving force
\newcommand{\emb}{x}			% embedding vector
\newcommand{\Cov}[1]{\mbox{Cov}(#1)}
\newcommand{\expec}[1]{E\left[ #1 \right]}
\newcommand{\eqref}[1]{Eq.~(\ref{#1})}
\newcommand{\Secref}[1]{Sec.~\ref{#1}}
\newcommand{\captionEM}[1]{\caption{\em #1}}
\newcommand{\OMIT}[1]{}

\title{%
\vspace{-0.5cm}{\small \tt \raggedleft{e-print published at
\href{http://arxiv.org/abs/0912.1064}{http://arxiv.org/abs/0912.1064} December 2009}} \\
\vspace{1.cm} 
On the numeric stability of the {S}{F}{A} implementation {\tt sfa-tk} \\
}
\author{Wolfgang Konen \\ \\
 Institute for Informatics, Cologne University of Applied Sciences \\
 Steinm\"ullerallee 1, D-51643 Gummersbach, Germany \\
 \texttt{\href{http://www.gm.fh-koeln.de/~konen}{http://www.gm.fh-koeln.de/{\footnotesize$\sim$}konen}} \\
 \texttt{\href{mailto:wolfgang.konen@fh-koeln.de}{wolfgang.konen@fh-koeln.de}}
}

\date{}

\begin{document} 

\maketitle

\begin{abstract}
Slow feature analysis (SFA) is a method for extracting slowly varying features from a quickly varying multidimensional signal. An open source \textsc{Matlab}-implementation {\tt sfa-tk} makes SFA easily useable. We show here that under certain circumstances, namely when the covariance matrix of the nonlinearly expanded data does not have full rank, this implementation runs into numerical instabilities. We propse a modified algorithm based on singular value decomposition (SVD) which is free of those instabilities even in the case where the rank of the matrix is only less than 10\% of its size. Furthermore we show that an alternative way of handling the numerical problems is to inject a small amount of noise into the multidimensional input signal which can restore a rank-deficient covariance matrix to full rank, however at the price of modifying the original data and the need for noise parameter tuning.
\end{abstract}

%PRL \keywords{driving force, nonlinear time series analysis, nonstationary time series, slow feature analysis}

\OMIT{
\section{Outline}
\begin{itemize}
	\item Intro
	\item Slow Feature Analysis (SFA) (OK)
	  \begin{itemize}
			\item The SFA approach (OK)
			\item show that $C = S C' S^T$ is $C=\Cov{\dot{z}}$ where $C'=\Cov{\dot{v}}$ (OK)
			\item Method GEN\_EIG: generalized eigenvalues \cite{Berkes2003} (OK)
			\item Method SVD\_SFA: from \cite{WisSej2002} (OK)
		\end{itemize}
	\item Improving {\tt sfa-tk} (OK)
	  \begin{itemize}
			\item 'broken' slow signal (OK)
			\item correct signal with Method SVD\_SFA (OK)
		\end{itemize}
	\item Discussion (OK)
	  \begin{itemize}
			\item general remarks (OK)
			\item noise injection (incl. a table with rank(B)) (OK)
			\item when do eigenvalues zero (or close to zero) appear? (OK)
			\item are the results dependent on the choice of $\epsilon$ (EV-cutoff) (OK)?
		\end{itemize}
	\item Conclusion (OK)
	\item Appendix A (OK)
	  \begin{itemize}
			\item covariance matrix, sphering matrix (OK)
			\item how to deal with zero eigenvalues (OK)
		\end{itemize}
	\item Appendix B (OK)
	  \begin{itemize}
			\item where new {\tt sfa-tk.V2} is available as open source (OK)
			\item new features (OK)
		\end{itemize}
	\end{itemize}
} % \OMIT

\section{Introduction} \label{sec:introduction}

Slow feature analysis (SFA) is an information processing method proposed by Wiskott and Sejnowski~\cite{WisSej2002} which allows to extract slowly varying signals from complex multidimensional time series. Wiskott~\cite{Wis98a} formulated a similar idea already before as a model of unsupervised learning of invariances in the visual system of vertebrates. SFA has been applied successfully to numerous different tasks: to reproduce a wide range of properties of complex cells
in primary visual cortex~\cite{BerWis2005}, to model the self-organized
formation of place cells in the hippocampus~\cite{FSW2007}, to classify handwritten digits~\cite{Berkes2005} and to extract driving forces from nonstationary time series~\cite{Wis2003c}.

The analysis of nonstationary time series plays an important role in the data understanding of various phenomena such as temperature drift in experimental setup, global warming in climate data or varying heart rate in cardiology. Such nonstationarities can be modeled by underlying parameters, referred to as
 driving forces, that change the dynamics of the system smoothly on a slow
 time scale or abruptly but rarely, e.g.\ if the dynamics switches between
 different discrete states.~\cite{Wis2003c}.  
 Often, e.g. in EEG-analysis or in monitoring of complex chemical or electrical power plants, one is particularly interested in revealing the driving forces themselves from the raw observed time series since they show interesting aspects of the underlying dynamics.
 %, for example the switching between different dynamic regimes.  
 
 We recently studied the notion of slowness in the context of driving force extraction from nonstationary time series~\cite{Kon09a}.
 There we used {\tt sfa-tk}~\cite{Berkes2003}, a freely available \textsc{Matlab}-implementation of SFA, to perform our experiments. These experiments requested to perform driving force extraction over a wide range of parameters. To our surprise in some seemingly very 'simple' time series which had a high regularity the original {\tt sfa-tk}-algorithm failed and produced 'slow' signals violating certain constraints as well as the slowness condition. We analyzed the problem and finally traced its source back to a rank deficiency in the expanded data's covariance matrix. In this paper we present this analysis and develop a new implementation which solves the problem.
 
 This paper is organized as follows: In \Secref{sec:SFA} we review the properties of the SFA approach and present the orignal {\tt sfa-tk}-algorithm as well as a slightly modified version which we will call SVD\_SFA. In \Secref{sec:Experiments} we show some driving force experiments demonstrating cases where the new SVD\_SFA algorithm gives good results while the original {\tt sfa-tk}-algorithm fails.
 \Secref{sec:discuss} discusses the properties of SVD\_SFA as well as an alternative approach based on noise injection.

\section{Slow Feature Analysis}
\label{sec:SFA}

 Slow feature analysis (SFA) has been originally developed in context of an
 abstract model of unsupervised learning of invariances in the visual
 system of vertebrates~\cite[]{Wis98a} and is described in detail
 in~\cite[]{WisSej2002,Wis2003c}.

\subsection{The SFA approach}
 We briefly review here the approach described in~\cite{Wis2003c}.
 The general objective of SFA is to extract slowly varying features from a
 quickly varying multidimensional signal.  
 
 Given is at time $t$ a raw input signal $\vec{s} = \vec{s}(t)$ as an  $m$-dimensional vector $\vec{s} = [s_1,...\s_m]$. A preprocessing step performs a sphering (see Appendix~A in \Secref{sec:AppendixA}) and an optional dimension reduction
\begin{equation}
	\vec{\is}(t) = \mat{W}_0\,(\vec{s}(t) - \expec{\vec{s}})
	\label{eq:inputSphere}
\end{equation}
 with $\expec{\cdot}$ indicating the temporal mean and with $\mat{W}_0$ as the $(n\times m)$ sphering matrix, $n \leq m$, whose $n$ rows are proportional to the $n$ largest eigenvectors of $\Cov{\vec{s}}$ (PCA). If $n=m$, no dimension reduction takes place. The preprocessed input $\vec{\is}(t) \in \Re^n$ is in any case mean-free and has a unit covariance matrix $\expec{\mat{x}\mat{x}^T}=\mat{1}$. (A simplified preprocessing which just normalizes each input component to mean 0 and variance 1 is also possible, but does not allow for dimension reduction.)
  
 For the preprocessed input $\vec{\is}(t)$ the SFA approach can be formalized 
 as follows:  Find the input-output function $\tf(\vec{\is})$ that generates a
 scalar output signal
\begin{equation}
 \os(t) := \tf(\vec{\is}(t))
\end{equation}
 with its temporal variation as  slowly as possible, measured by the variance
 of the time derivative:
\begin{equation}
 \mbox{minimize} \quad \Delta(y) = \expec{\dot{\os}^2} \label{eq:slowness}
\end{equation}
To avoid the trivial constant solution the output signal  has to meet the following constraints:
\begin{eqnarray}
 \expec{\os} &=& 0 \quad \mbox{(zero mean)} \,, \label{eq:constr0} \\
 \expec{\os^2} &=& 1 \quad \mbox{(unit variance)} \label{eq:constr1} \,.
\end{eqnarray}

 This is an optimization problem of variational calculus and as such
 difficult to solve. But if we constrain the input-output function to
 be a linear combination of some fixed and possibly nonlinear basis
 functions, the problem becomes tractable. 
 %%%%%%%%%%%%%%%%%%%%%%%%
 
  Let $\vtfc = \vtfc(\vec{\is}) \in \Re^M$ be a vector of some fixed basis
 functions.  To be concrete assume that $\vtfc$ contains all monomials of
 degree~1 and~2.  Applying $\vtfc$ to the input signal $\vec{\is}(t)$
 yields the nonlinearly expanded signal $\vec{\tfc}(t)$:
%
%PRL \begin{subequations}
\begin{eqnarray}
 \vtfc(\vec{\is}) &:=& [\is_1,\, \is_2,\, ...,\, \is_n,\, 
 \is_1^2,\, \is_1 \is_2,\, ...,\, \is_n^2]^T, \label{eq:expand} \\ 
 \vec{\tfc}(t) &:=& \vtfc(\vec{\is}(t)) \,.
\end{eqnarray}
%PRL \end{subequations}

 For reasons that become clear below, it is convenient to sphere (or
 whiten) the expanded signal and transform the basis functions accordingly:
%
%PRL \begin{subequations}
\begin{eqnarray}
 \zes(\vec{\is}) &:=& \mat{S} \ (\vtfc(\vec{\is}) - \expec{
 \vec{\tfc} }) \,, \\
 \vec{\es}(t) &:=& \mat{S} \ (\vec{\tfc}(t) - \expec{ \vec{\tfc}
 }) \,,
\end{eqnarray}
%PRL \end{subequations}
%
 with the sphering matrix $\mat{S}$ (see Appendix~A in\Secref{sec:AppendixA}) chosen such that the signal components
 have a unit covariance matrix, i.e. 
 \begin{equation}
	\mat{1} = \Cov{\vec{z}} = \expec{ \vec{\es} \vec{\es}^T } = 
	\mat{S}\,\Cov{\vec{v}}\,\mat{S}^T \,,
	\label{eq:Covz}
\end{equation}
which can be done easily with the help of singular value decomposition (SVD). However, such a sphering matrix $\mat{S}$ exists only, if $\mat{B} := \Cov{\vec{v}}$ has no eigenvalues exactly zero (or close to zero).   
 The signal components have also zero mean, $\expec{ \vec{\es} } = \vec{0}$, 
 since the mean values have been subtracted.
 
 %\vspace{1em}
 \paragraph{Proposition 1} Let 
$
	\mat{C} := \expec{ \dot{\vec{\es}} \dot{\vec{\es}}^T }
$
be the time-derivative covariance matrix of the sphered signals. Calculate with SVD the eigenvectors $\vec{w}_j$ of $\mat{C}$ with their eigenvalues $\lambda_1<\lambda_2<\lambda_3<...$, i.e.
\begin{equation}
	\mat{C}\vec{w}_j = \lambda_j\vec{w}_j
	\label{eq:CEigen}
\end{equation}
(we assume here that the eigenvalues are pairwise distinct and that the eigenvectors have norm 1, i.e. $\vec{w}_j^T \vec{w}_k = \delta_{jk}$).
Then the output signals defined as 
\begin{equation}
	\os_j(t)=\vec{w}_j^T \vec{z(t)}	
	\label{eq:w_j}
\end{equation}
have the desired properties: 
\begin{eqnarray}
 \expec{\os_j} &=& 0 \quad \mbox{(zero mean)} \,, \nonumber \\
 \expec{\os_j^2} &=& 1 \quad \mbox{(unit variance)} \,, \nonumber \\
 \expec{\os_j \os_k} &=& 0 \quad \mbox{(decorrelation)}\,, j\neq k \,, \label{eq:decor}\\
 \expec{\dot{\os_j}^2} &=& \lambda_j\,. \nonumber
\end{eqnarray}

\paragraph{Proof}  The first property is obvious since $\expec{ \vec{\es} } = \vec{0}$, the second and third property are direct consequences of the sphered signal and the orthonormality of the eigenvectors: 
$$  
	\expec{\os_j \os_k} = \expec{\vec{w}_j^T \vec{z(t)} \vec{z(t)}^T\vec{w}_k} =
	\vec{w}_j^T \underbrace{\expec{ \vec{\es} \vec{\es}^T }}_{= \mat{1}} \vec{w}_k = \delta_{jk}
$$	
Finally the fourth property comes from the eigenvalue equation:
$$
	\expec{\dot{\os_j}^2} = \expec{\vec{w}_j^T \dot{\vec{z}}(t) \dot{\vec{z}}(t)^T\vec{w}_j} =
	\vec{w}_j^T \expec{\dot{\vec{z}}\dot{\vec{z}}^T} \vec{w}_j =
	\vec{w}_j^T \mat{C} \vec{w}_j = \vec{w}_j^T \lambda_j \vec{w}_j = \lambda_j\,. \spadesuit
$$

Thus the sequence $\os_1(t)\,,\os_2(t), ...$ constitutes a series of slowest, next-slowest, next-next-slowest, ... signals,  where each signal is completely decorrelated to all preceeding signals and each signal is the slowest signal among those decorrelated to the preceeding ones.  

\paragraph{Proposition 2} If $\mat{B}:=\Cov{\vec{v}}$ is regular (i.e. it has no zero eigenvalues and thus $\mat{S}$ and $\mat{S}^{-1}$ exist) then (and only then) \eqref{eq:CEigen} is equivalent to the generalized eigenvalue equation of Berkes~\cite{Berkes2003}: 
\begin{equation}
	\mat{C'}\,\vec{w}_j'=\lambda_j \mat{B}\,\vec{w}_j' \qquad \mbox{with} \qquad  
	\mat{C'} := \expec{ \dot{\vec{v}} \dot{\vec{v}}^T }
	\label{eq:genEigen}
\end{equation}
 and both eigenvalue equations have the same eigenvalues $\lambda_j$.

\paragraph{Proof} With $\dot{\vec{z}} = \mat{S}\dot{\vec{v}}$ we have 
\begin{equation}
	\mat{C} = \expec{ \dot{\vec{\es}} \dot{\vec{\es}}^T }
				  = \mat{S} \expec{ \dot{\vec{v}} \dot{\vec{v}}^T } \mat{S}^T
				  = \mat{S}\mat{C'}\mat{S}^T 
	\label{eq:CwithS}
\end{equation}
Using this and the identity \eqref{eq:S} from the Appendix~A in \Secref{sec:AppendixA} we can rewrite \eqref{eq:CEigen}
\begin{eqnarray}
	\mat{C}\vec{w}_j &=& \lambda_j\vec{w}_j \nonumber\\
	\mat{S}\mat{C'}\mat{S}^T \vec{w}_j &=& 
	\lambda_j \mat{S}\, \mat{B} \,\underbrace{\mat{S}^T\vec{w}_j}_{=\vec{w}_j'} \nonumber\\
	\mat{C'}\,\vec{w}_j' &=& \lambda_j \, \mat{B}\,\vec{w}_j' \nonumber
\end{eqnarray}
where we have multiplied with $\mat{S}^{-1}$ from the left in the third line. $\spadesuit$
%%%
%%% TODO: can we show that gen-eig fails, if B is singular??
%%%

%%%%%%%%%%%%%%%%%%%%%%%%%%%%%%%%%%%%%%%%

 Basically, SFA consists of the following four steps: (i)~expand the input
 signal with some set of fixed possibly nonlinear functions; (ii)~sphere
 the expanded signal to obtain components with zero mean and unit
 covariance matrix; (iii)~compute the time derivative of the sphered
 expanded signal and determine the normalized eigenvectors of its covariance
 matrix; (iv)~project the sphered expanded
 signal onto this eigenvectors to obtain the output signals $y_j$.
 In the following Sections~\ref{sec:GEN_EIG} and \ref{sec:SVD_SFA} we present two different ways of implementing the SFA approach.

\subsection{Algorithm GEN\_EIG}
\label{sec:GEN_EIG}
Algorithm GEN\_EIG is based on Proposition 2 and is the SFA algorithm as implemented in {\tt sfa-tk}~\cite{Berkes2003} and as described in \cite{Berkes2005}.
Given is at time $t$ a raw input signal $\vec{s}(t)\in \Re^m$. Let $n$ be the dimension of the preprocessed input $\vec{x}(t)$ and $M$ be the dimension of the expanded signal $\vec{v}(t)$ (e.g. with monomials of degree 2). 

\paragraph{Unsupervised training on training signal $\vec{s}(t)$}
\begin{algorithmic}[1]    % [1] bedeutet: Nummeriere jede Zeile, ohne "[ ]": keine Zeilennummern
  \State $\vec{\is}(t) := \mat{W}_0\,(\vec{s}(t) - \vec{s}_0) \in \Re^n, 
  				\quad \vec{s}_0:=\expec{\vec{s}}$
  			\Comment{Preprocess the raw input signal (sphering, PCA), \eqref{eq:inputSphere}}
	\State $\vec{\tfc}(t) := \vtfc(\vec{\is}(t)) \in \Re^M, \quad \vec{v}_0:=\expec{\vec{v}}$
				\Comment{Expand the preprocessed signal, see for example \eqref{eq:expand}}
	\State $\mat{B} := \Cov{\vec{v}}$ 	\Comment{Covariance matrix of expanded data, \eqref{eq:covmat}} 
	\State $\mat{C'} := \expec{ \dot{\vec{v}} \dot{\vec{v}}^T }$ 
				\Comment{Time-derivative ($\dot{v}$) covariance matrix, \eqref{eq:genEigen}}
	\State Solve $\mat{C'}\,\vec{w}_j'=\lambda_j \mat{B}\,\vec{w}_j'$		for $\{ \lambda_j, \vec{w}_j'\}	$	
				\Comment{Generalized eigenvalue equation, \eqref{eq:genEigen}}
  \State Return $\{\mat{W}_0, \vec{s}_0, \vec{v}_0, \lambda_j, \vec{w}_j' \,|\, j=1,\ldots,M\}$
\end{algorithmic}
\vspace{1em}
The results returned from the training step can be applied to any input signal $\vec{s}(t)$ in the following way to yield the slow signals $y_j(t)$:
 
\paragraph{Application to (training or new) signal $\vec{s}(t)$}
\begin{algorithmic}[1]    % [1] bedeutet: Nummeriere jede Zeile, ohne "[ ]": keine Zeilennummern
  \State $\vec{\is}(t) = \mat{W}_0\,(\vec{s}(t) - \vec{s}_0) \in \Re^n$
  			\Comment{Preprocess the raw input signal (sphering, PCA), \eqref{eq:inputSphere}}
	\State $\vec{\tfc}(t) = \vtfc(\vec{\is}(t)) \in \Re^M$
				\Comment{Expand the preprocessed signal, see for example \eqref{eq:expand}}
	\State $y_j(t) = \vec{w}_j'^T\, (\vec{v}(t)-\vec{v}_0) \in \Re, \quad  j=1,\ldots,M$
				\Comment{$y_1(t)$: slowest, $y_2(t)$: 2nd slowest signal and so on}
  \State Return $\{y_j(t) \,|\, j=1, \ldots, M \}$
\end{algorithmic}

\vspace{1em}
Algorithm GEN\_EIG works fine as long as $\mat{B}$ does not contain zero (or close to zero) eigenvalues. But if $\mat{B}$ becomes singular (or close to singular) then Proposition 2 does no longer hold and we will see in \Secref{sec:Experiments} that the slow signals $y_j$ may become wrong. Therefore we reformulated the SFA approach in such a fashion that we can control (sort out) zero (or very small) eigenvalues and present the result as algorithm SVD\_SFA.

\subsection{Algorithm SVD\_SFA}
\label{sec:SVD_SFA}

 Algorithm SVD\_SFA is based on \cite{WisSej2002} and Proposition 1. It is implemented as an extension of {\tt sfa-tk}.
 The application part of SVD\_SFA is the same as in algorithm GEN\_EIG. The unsupervised training part is the same for Steps 1.-4., but somewhat different from Step 5. on: 

\paragraph{Unsupervised training on training signal $\vec{s}(t)$}
\begin{algorithmic}[1]    % [1] bedeutet: Nummeriere jede Zeile, ohne "[ ]": keine Zeilennummern
  \State $\vec{\is}(t) := \mat{W}_0\,(\vec{s}(t) - \vec{s}_0) \in \Re^n, 
  				\quad \vec{s}_0:=\expec{\vec{s}}$
  			\Comment{Preprocess the raw input signal (sphering, PCA), \eqref{eq:inputSphere}}
	\State $\vec{\tfc}(t) := \vtfc(\vec{\is}(t)) \in \Re^M, \quad \vec{v}_0:=\expec{\vec{v}}$
				\Comment{Expand the preprocessed signal, see for example \eqref{eq:expand}}
	\State $\mat{B} := \Cov{\vec{v}}$ 	\Comment{Covariance matrix of expanded data, \eqref{eq:covmat}} 
	\State $\mat{C'} := \expec{ \dot{\vec{v}} \dot{\vec{v}}^T }$ 
				\Comment{Time-derivative ($\dot{v}$) covariance matrix, \eqref{eq:genEigen}}
	\State Find $\mat{S}$ such that $ \mat{S}\mat{B}\mat{S}^T = \mat{1} $	
				\Comment{Sphere expanded $\vec{\tfc}$ with SVD, see \eqref{eq:sphere}, \eqref{eq:sphere0}} 
	\State $ \mat{C} := \mat{S}\mat{C'}\mat{S}^T $
				\Comment{Time-derivative ($\dot{z}$) covariance matrix, \eqref{eq:CwithS}}
	\State Solve $\mat{C}\,\vec{w}_j=\lambda_j \,\vec{w}_j$		for $\{ \lambda_j, \vec{w}_j\}	$	
				\Comment{Eigenvalue equation, \eqref{eq:CEigen}}
	\State $\vec{w}_j':=\mat{S}^T\,\vec{w}_j \in \Re^P$
  \State Return $\{\mat{W}_0, \vec{s}_0, \vec{v}_0, \lambda_j, \vec{w}_j' \,|\, j=1,\ldots,P\}$
\end{algorithmic}

\vspace{1em}
 The main difference is Step 5. where the sphering matrix $\mat{S}$ for $\mat{B} := \Cov{\vec{v}}$ is calculated with SVD according to the lines described in Appendix~A (\Secref{sec:AppendixA}). This works even in the case that $\mat{B}$ is singular. In that case some $M-P$ rows of $\mat{S}$ are zero (more precisely: each row whose eigenvalue fulfills the 'close-to-zero' condition of~\eqref{eq:closeToZero}). These rows are either directly excluded from $\mat{S}$, making it a ($P \times M$) matrix, or they are kept and lead to $M-P$ eigenvalues $\lambda_j$ of $\mat{C}$ which are exactly zero and which are then excluded from further analysis.
 In the case where $\mat{B}$ is regular, we have $P=M$ and no eigenvalue is zero. In any case, algorithm SVD\_SFA will return $P$ non-zero eigenvalues, $P \leq M$, and their corresponding eigenvectors.
 
 In Step 6. of the unsupervised training we calculate $\mat{C}=\expec{ \dot{\vec{\es}} \dot{\vec{\es}}^T}$ without the need for calculating $\vec{\es}$ explicitly. Likewise, Step 8. prepares 
 $\vec{w}_j'$ in such a way that 
$$
	\vec{w}_j'^T\, (\vec{v}(t)-\vec{v}_0) = \vec{w}_j^T\, \mat{S}\, (\vec{v}(t)-\vec{v}_0) = 
	\vec{w}_j^T \vec{z}(t)
$$ 
so that Step 3. of the application part leads to the output signal as requested by Proposition 1.  

Why does the sphered expanded signal $\vec{z}(t)$, which is central to Proposition 1, not appear in algorithm SVD\_SFA? The algorithm is deliberately implemented in such a way that direct access to $\vec{z}(t)$ is not necessary. The reason is that for large input sequences the implementation of {\tt sfa-tk} allows to pass over the input data chunk-by-chunk into the expansion steps (2.-4.), where $\vec{v}, \, \mat{B}$ and $\mat{C'}$ are formed, which is considerably less memory-consuming. But only when $\mat{B}$ is established at the end of the expansion phase, we can calculate $\mat{S}$. If a direct representation of $\vec{z}(t) = \mat{S}\,(\vec{v}(t)-\vec{v}_0)$ were necessary, another chunkwise pass-over of the input would be necessary and considerably slow down the computation. With the indirect method in Steps~6. and 8. we accomplish the same result without the need for $\vec{z}(t)$. 

\begin{figure*}[tbp]

%\centerline{\includegraphics[width=0.926\textwidth]{figure2a}\hspace*{0.044\textwidth}}
\centerline{\includegraphics[width=0.49\textwidth]{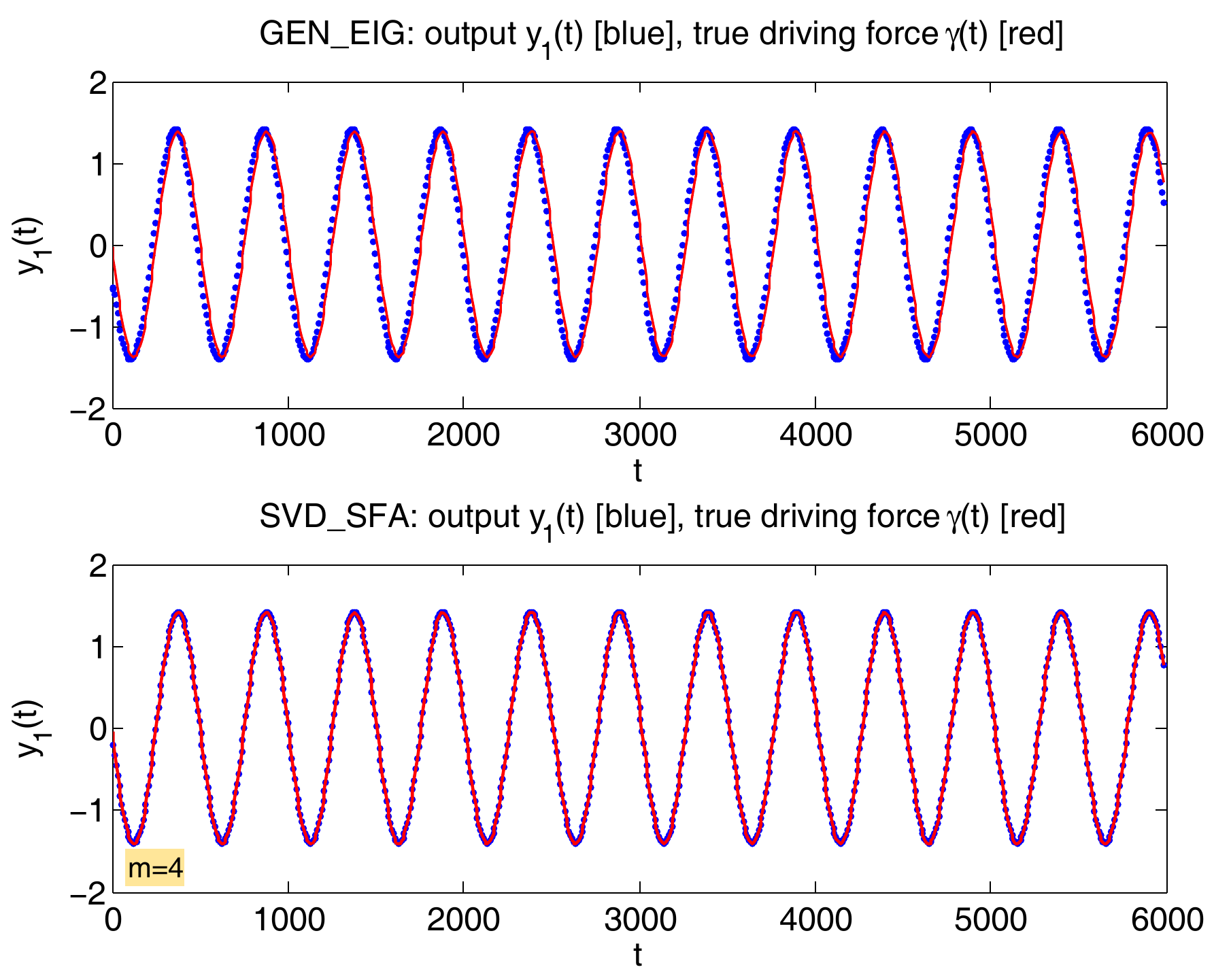}
						\includegraphics[width=0.49\textwidth]{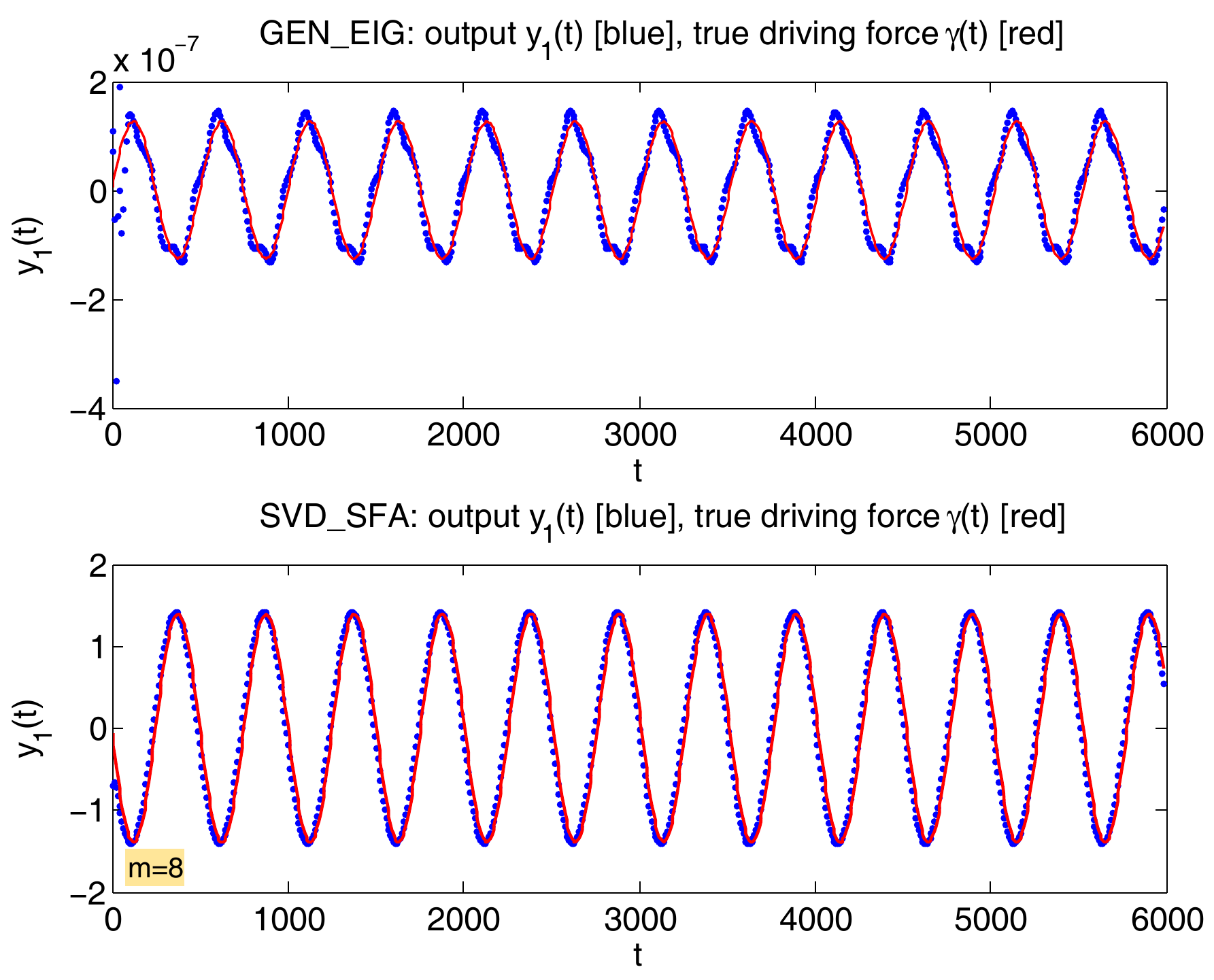}}
%%%%
%%%% generated with script sfa-tk/demo/arxiv2009_SFA2.m using tau=1;nuf=25;q=1.2 which writes  
%%%% onto images/eps/. Go there and convert drive-q1.2-m04-nuf25.eps and drive-q1.2-m08-nuf25.eps 
%%%% with Adobe Distiller (when saving, replace 'q1.2' with 'q1p2' since LaTeX does not like
%%%% two points in filenames).
%%%%
 \captionEM{\label{fig:GENvsSVD1}  Each diagram shows the SFA-output $y_1(t)$ (blue dots) and the aligned true driving force  $A(\p(t);y(t))$ (red line) (see \eqref{eq:align1}). The SFA-output was generated from a logistic map time series $\uu(t)$ (\eqref{eq:logistic2}) with parameters $q=1.2,\,\, \tau=1$ and the driving force of \eqref{eq:gamma}. The upper row shows the GEN\_EIG implementation and the lower row the SVD\_SFA implementation. Left: $m=4$, right: $m=8$. We see that already at embedding dimension $m=8$ the GEN\_EIG results starts to deteriorate, since it has only an amplitude in the order of $10^{-7}$.}
\end{figure*}

\begin{figure*}[htbp]
\centerline{\includegraphics[width=0.49\textwidth]{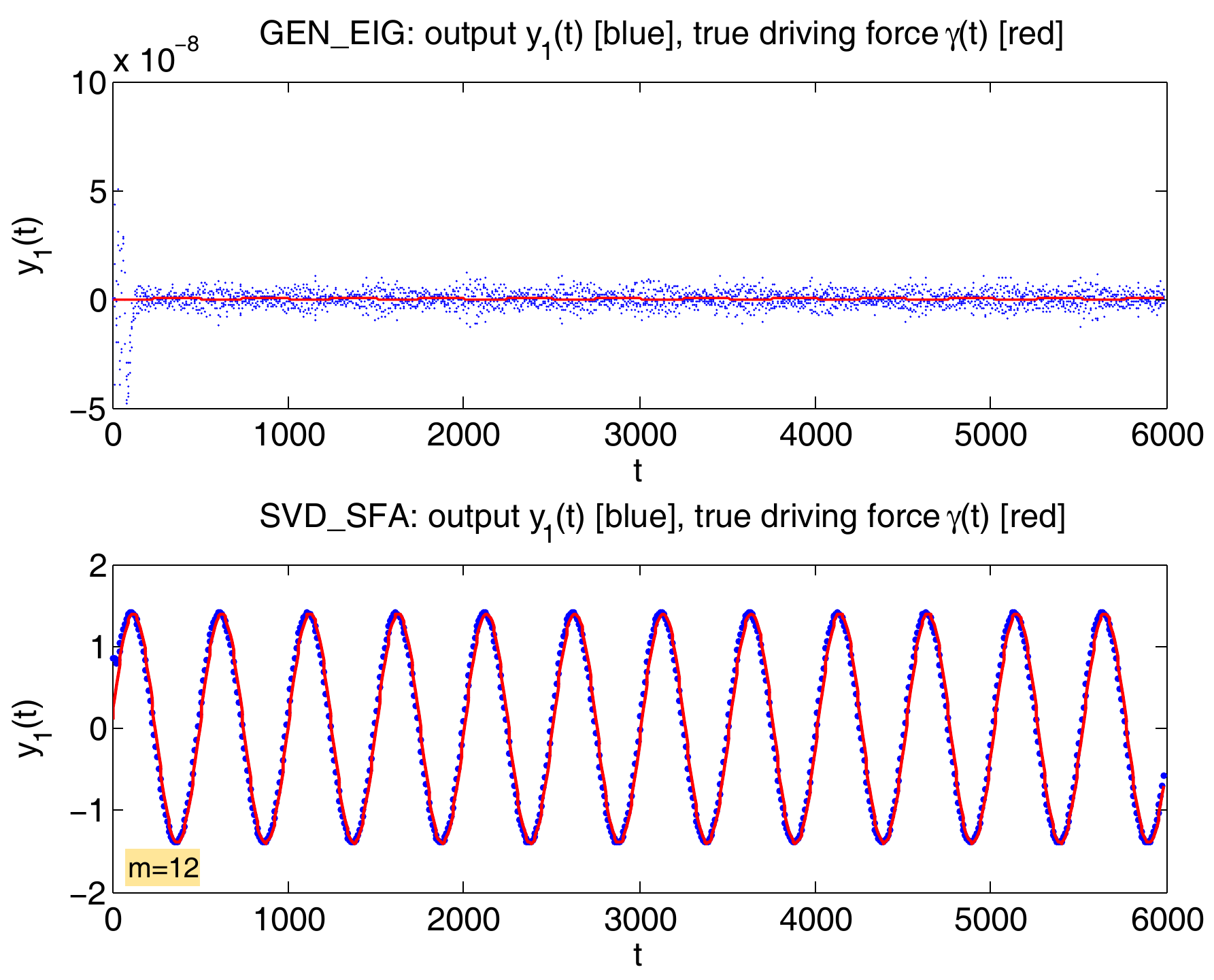}
						\includegraphics[width=0.49\textwidth]{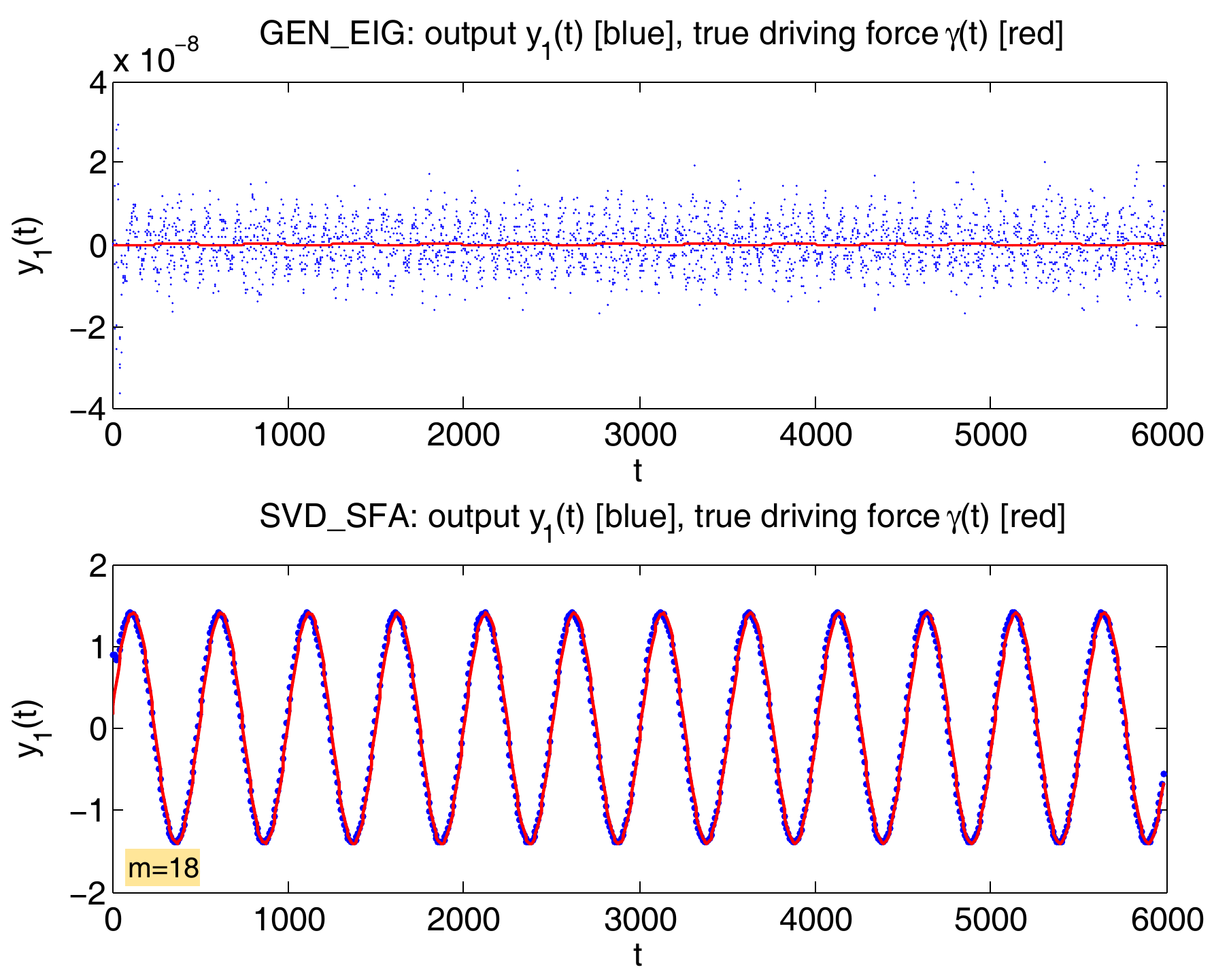}}
%%%%
%%%% generated with script sfa-tk/demo/arxiv2009_SFA2.m using tau=1;nuf=25;q=1.2 which writes  
%%%% onto images/eps/. Go there and convert drive-q1.2-m12-nuf25.eps and drive-q1.2-m18-nuf25.eps 
%%%% with Adobe Distiller (when saving, replace 'q1.2' with 'q1p2' since LaTeX does not like
%%%% two points in filenames).
%%%%
 \captionEM{\label{fig:GENvsSVD2} Same as  Fig.~\ref{fig:GENvsSVD1}, only for larger $m=12$ and $m=18$. Now the results from GEN\_EIG are completely deteriorated, they are by no means slow nor do they have unit variance, while SVD\_SFA still gives good results.}
\end{figure*}

\vspace{1em}
 In the rest of this paper we will work with time series $\vec{\is}(t_i)$ instead of
 continuous signals $\vec{\is}(t)$, but the transfer of the algorithms
 described above to time series is straight forward.  The time derivative
 is simply computed as the difference between successive data points
 assuming a constant sampling spacing $\Delta t = 1$.

\section{Experiments}
\label{sec:Experiments}
 In the following we present examples with time series $\uu(t)$
 derived from a logistic map to illustrate the results with different implementations of
 SFA. The underlying driving force is denoted by $\p(t)$ and may vary
 between $-1$ and $1$ smoothly and considerably slower (as defined by the variance of its time derivative
 (\ref{eq:slowness})) than the  time series $\uu(t)$. The approach follows closely the work of Wiskott~\cite{Wis2003c} and own previous work~\cite{Kon09a}, using a very simple driving force in the shape of just one sinusoidal frequency component: 
\begin{equation}
	\p(t) = \sin(0.0125 t) \quad\in\quad [-1, 1]
	\label{eq:gamma}
\end{equation} 
The logistic map is formulated with a parameter $q \in [0.1,3.9]$ allowing to control the presence or absence of chaotic motion (see~\cite{Kon09a} for details): 
\begin{equation}
  \uu(t+1) = (4.0-q+0.1 \p(t)) \uu(t) (1-\uu(t)) \,,
  \label{eq:logistic2}
\end{equation}

 Taking the time series $\uu(t)$ directly as an input signal would not give
 SFA enough information to estimate the driving force, because SFA
 considers only data (and its derivative) from one time point at a time.
 Thus it is necessary to generate an embedding-vector time series as an
 input.  Here embedding vectors at time point $t$ with delay $\tau$ and
 dimension $m$ are defined by
\begin{eqnarray}
 \vec{\emb}(t) &:=& [	\uu(t - \tau (m-1)/2),\, 
 											\uu(t - \tau ((m-1)/2-1)),\, ..., 
 											\uu(t + \tau (m-1)/2)]^T \,,
\end{eqnarray}
 for scalar $\uu(t)$ and odd $m$.  The definition can be easily extended to
 even $m$, which requires an extra shift of the indices by $\tau/2$ or its
 next lower integer to center the used data points at $t$.  Centering the
 embedding vectors results in an optimal temporal alignment between
 estimated and true driving force.

In order to inspect visually the agreement between a slow SFA-signal and the driving force $\p(t)$ we must bring the driving force into alignment with $y(t)$ (the scale and offset of slow signals $y(t)$ extracted by SFA are arbitrarily fixed by the constraints and the sign is random). Therefore we define an {\em $y$-aligned signal}
\begin{equation}
	A(\p(t);y(t)) = a\p(t)+b
	\label{eq:align1}
\end{equation}
where the constants $a$ and $b$ are chosen in such a way that 
\begin{equation}
	\expec{ (a\p(t)+b - y(t))^2 } \stackrel{!}{=} Min.
	\label{eq:align2}
\end{equation}

\begin{table}
	\label{tab:GENvsSVD}
	\captionEM{ Results from the implementation experiments where index $G$ denotes GEN\_EIG and index $S$ denotes SVD\_SFA. $m$: embedding dimension, $N_G$: number of eigenvalues of the generalized eigenvalue equation, $N_S$: number of eigenvalues of $\Cov{\dot{\vec{z}}}$, the constraints $\expec{y_1}$ and $\expec{y_1^2}$ acc. to \eqref{eq:constr0}, \eqref{eq:constr1} and the slowness indicator $\eta$~\cite{WisSej2002,Kon09a}. Low $\eta$-values indicate slow signals, high $\eta$-values fast signals.
	For algorithm GEN\_EIG and embedding dimension $m\geq 8$ neither the unit variance constraint $\expec{y_1^2}_G=1$ nor the slowness principle for $\eta_G$ are fulfilled. 
	}
\begin{center}
	\begin{tabular}{|c|r|r|r|r|r|r|r|r|}
	\phline
$m$ &$N_G$&$N_S$&$\expec{y_1}_G$&$\expec{y_1}_S$&
													   $\expec{y_1^2}_G$& $\expec{y_1^2}_S$ & $\eta_G$& $\eta_S$ \\ \phline\hline
	2	&   5&	5	&	 1.1e-16	&	-4.8e-18&	1			 &	1& 	  11.8&	11.83 \\ \hline
	4	&  14&	9	&	-7.2e-16	&	 9.0e-17&	1			 &	1&	  11.8&	11.82 \\ \hline
	8	&  44&	17&	-1.8e-16	&	 6.9e-17&	9.5e-08&	1&	 387.7&	11.79 \\ \hline
%	9	&  54&	20&	-2.3e-16	&	 2.1e-18&	1.2e-08&	1&	 612.7&	11.78 \\ \hline
	10&  65&	22&	-3.2e-17	&	 2.5e-16&	1.0e-07&	1&	  96.0&	11.77 \\ \hline
% 11&  77&	23&	-8.0e-17	&	-3.7e-18& 1.9e-08&	1&	 513.0&	11.77 \\ \hline
  12&  90&	23&	 1.9e-16	&	 8.8e-17&	5.6e-09&	1&	1572.1&	11.77 \\ \hline
% 16& 152&	25&	 1.3e-17	&	 2.1e-16&	9.2e-09&	1&	1750.6&	11.77 \\ \hline
% 18& 189&	25&	-4.0e-17	&	-7.4e-17&	5.8e-09&	1&	971.25&	11.77 \\ \hline
  20& 230&	26&	 1.2e-16	&	-7.3e-17&	6.0e-09&	1&	1480.3&	11.76 \\ \hline
% 25& 350&	26&	-2.2e-17	&	-2.0e-16&	8.1e-09&	1&	1548.1&	11.79 \\ \hline
  30& 495&	26&	 1.6e-16	&	 1.5e-17&	4.7e-09&	1&	1661.6&	11.76 \\ \hline
	\end{tabular}	
	%%%%
	%%%% generated as array ewarr from sfa-tk/demo/arxiv2009_SFA2.m using tau=1;nuf=25;q=1.2 
	%%%%
\end{center}
\end{table}

 The following simulations are based on 6000 data points each and were done
 with \textsc{Matlab 7.0.1} (Release 14) as well as with \textsc{Matlab 7.6.0} (R2008a) using the SFA toolkit {\tt sfa-tk}~\cite{Berkes2003} and our extensions to it.
 
 Fig.~\ref{fig:GENvsSVD1} and Fig.~\ref{fig:GENvsSVD2} show the results for some embedding dimensions in the case of the logistic map with $q=1.2$. The GEN\_EIG results in Fig.~\ref{fig:GENvsSVD2} are completely corrupted by numerical errors and do not show anything 'slow'. Table~\ref{tab:GENvsSVD} shows indicative numbers for the same experiment. The number of eigenvalues $N_G$ is always equal to the embedding dimension $M$ which is for $m\geq 8$ much larger than the true dimensionality of the expanded data. In contrast, the number $N_S$ of (non-zero) eigenvalues returned from SVD\_SFA approaches a limiting value, in this case 26, as $m$ increases. The mean value $\expec{y_1}$ is always correct since $y_1$ is built from mean-free components (cf. \eqref{eq:w_j}). The unit-variance constraint $\expec{y_1^2}_G=1$ is violated for $m\geq 8$ in the case of GEN\_EIG. Likewise, the slowness indicator $\eta_G$ becomes orders of magnitude larger than $11.8$ which is the slowness of the true driving force (i.e. the signal is very fast).

\begin{figure*}[htbp]
\centerline{\includegraphics[width=0.59\textwidth]{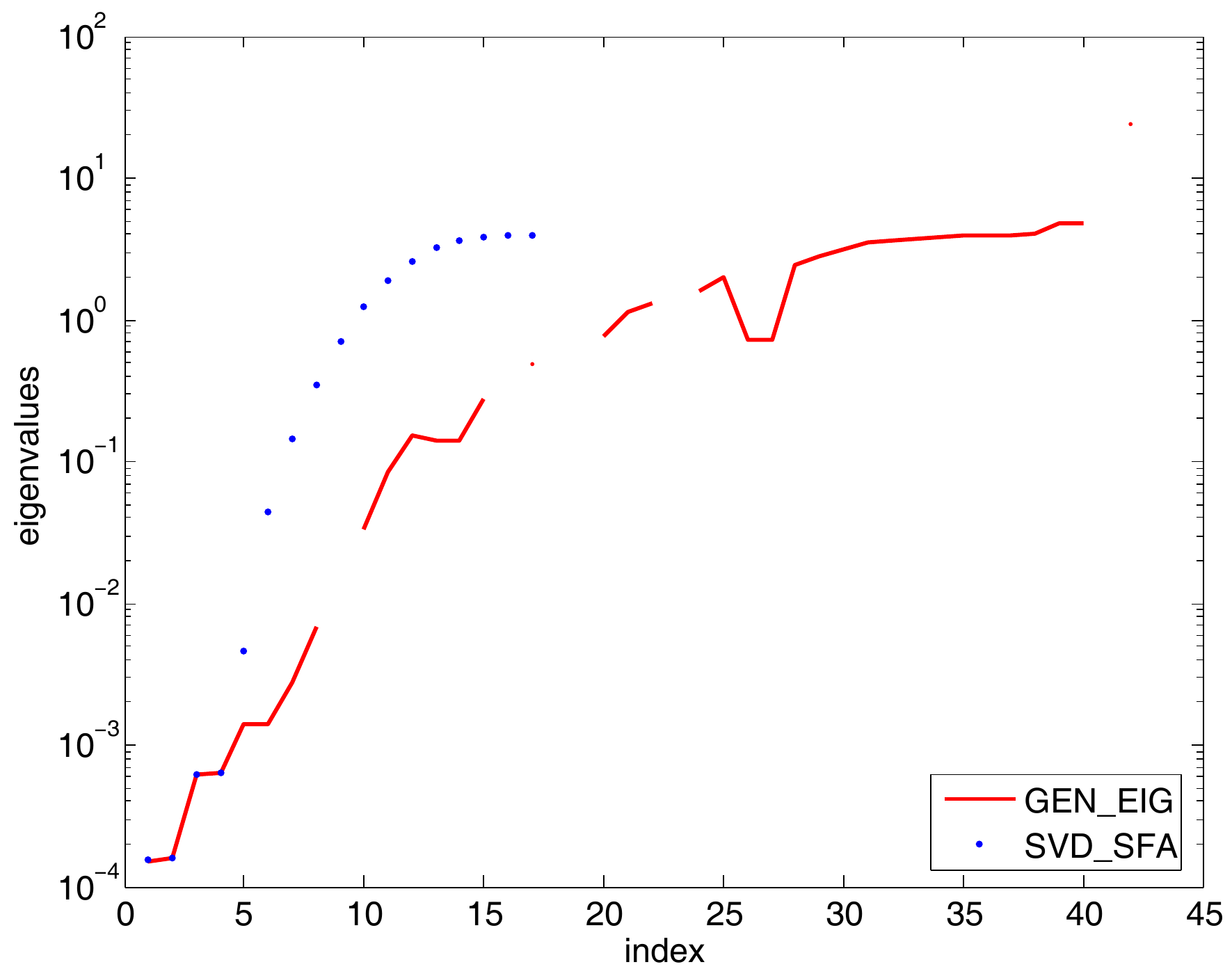}}
%%%%
%%%% generated with script sfa-tk/demo/arxiv2009_SFA2.m using tau=1;nuf=25;q=1.2 which writes  
%%%% onto images/eps/. Go there and convert drive-q1.2-m08-nuf25.eps and drive-q1.2-m18-nuf25.eps 
%%%% with Adobe Distiller (when saving, replace 'q1.2' with 'q1p2' since LaTeX does not like
%%%% two points in filenames.
%%%%
 \captionEM{\label{fig:eigenval} Eigenvalues $\lambda_j$ for both algorithms GEN\_EIG (red line) and SVD\_SFA (blue dots) for the settings of Fig.~\ref{fig:GENvsSVD1} and embedding dimension $m=8$, where $\mat{B}$ is singular. The red line is broken whenever a negative or complex eigenvalue occurs for GEN\_EIG.}
\end{figure*}

\section{Discussion}
\label{sec:discuss}
We can summarize the above experiments as follows: the standard implementation GEN\_EIG of SFA in {\tt sfa-tk} is likely to fail if $\mat{B}$, the covariance matrix of the expanded data $\vec{v}$, becomes singular. The fact that $\mat{B}$ becomes singular indicates that the dimension of the expanded space is higher than the 'true' dimensionality of the data. The failure can be traced back to \textsc{Matlab}'s routine {\tt eig(A,B)} for calculating generalized eigenvalues, which is said according to \textsc{Matlab}'s Help to work also in the case of singular $\mat{B}$ but apparently is not. 

We show in Fig.~\ref{fig:eigenval} the eigenvalues $\lambda_j$ for both algorithms GEN\_EIG and SVD\_SFA for one example with singular $\mat{B}$ (embedding dimension $m=8$). It is seen that in the case of GEN\_EIG the line of eigenvalues is interrupted several times which is due to the fact that the corresponding eigenvalues are either negative or complex and can not be shown on a logarithmic scale. Clearly negative or complex eigenvalues should not occur for symmetric matrices $\mat{B}$ and $\mat{C'}$. If we use SVD (command {\tt svd} in \textsc{Matlab}) to calculate the eigenvalues of the singular, symmetric matrices $\mat{B}$ and $\mat{C}$ we find only real, nonnegative eigenvalues, as it should be. 

\paragraph{Does a singular matrix $\mat{B}$ frequently occur?} 

A singular matrix $\mat{B}$ is only likely to occur if the data show a high regularity as for example in the synthesized time series of our experiments above. For data from natural sources or data with a certain amount of noise a singular $\mat{B}$ is not likely to happen, at least if the length of the time scale is not shorter than the embedding dimension. Even in our experiments with the synthesized logistic map, if we move to the chaotic region $q\leq 0.4$ then no singular $\mat{B}$ is observed, not for low and not for high embedding dimensions $m$, and the GEN\_EIG algorithm of {\tt sfa-tk} works as well as SVD\_SFA.

Thus for natural data or for data with noise it is very unlikely to see a singular $\mat{B}$ and this is perhaps the reason why the weakness of the GEN\_EIG algorithm remained so far unnoticed. 

But in certain circumstances (regular data or small amount of data, as it may occur more frequently in classification applications with a limited number of patterns per class) a singular $\mat{B}$ can nevertheless happen and it is good to have with SVD\_SFA a numerically stable approach. 

\begin{table}[btp]
{	
	\captionEM{Rank of matrix $\mat{B}=\Cov{\vec{v}}$ as a function of embedding dimen\-sion $m$ and noise injection ampli\-tude $\sigma_{\nu}$. $M$ is the size of the expanded signal $\vec{v}$ which equals the number of rows and columns of $\mat{B}$. For each noise amplitude lower than $\sigma_{\nu}=10^{-4}$ certain rank deficiencies occur. 
	\label{tab:rank}}
\begin{center}
	\begin{tabular}{|r|r||r|r|r|r|r|r|}
	\hline	
	 & $\sigma_{\nu}$
   &   0& $10^{-10}$& $10^{-8}$& $10^{-6}$& $10^{-4}$ &  \\ \phline
$m$ & & \multicolumn{5}{c|}{$\mbox{rank}(\mat{B})$}& $M$ \\ \phline\hline
 2&&	 5&	  5&	  5&	  5&	  5&	  5 \\ \hline
 4&&	14&	 14&	 14&	 14&	 14&	 14 \\ \hline
 8&&	24&	 28&	 31&	 43&	 44&	 44 \\ \hline
10&&	30&	 39&	 48&	 64&	 65&	 65 \\ \hline
12&&	32&	 54& 	 69& 	 88&	 90&	 90 \\ \hline
20&&	35&	141&	189&	227&	230&	230 \\ \hline
30&&	35&	236&	406&	490&	495&	495 \\ \hline
	\end{tabular}	
	%%%%
	%%%% generated as array 'ranks' from sfa-tk/demo/arxiv2009_SFA2_rank.m 
	%%%% (using tau=1;nuf=25;q=1.2 in arxiv2009_SFA2.m)
	%%%%
\end{center}
}
\end{table}
\begin{table}[hbtp]
{
	\captionEM{Standard deviation $\expec{y_1^2}$ of the slowest signal from GEN\_EIG as a function of embedding dimen\-sion $m$ and noise injection amplitude $\sigma_{\nu}$. We see that even in a case where the rank-deficiency of $\mat{B}$ is only 1 out of 65 dimensions (e.~g. for $m=10, \sigma_{\nu}=10^{-6}$, see Tab.~\ref{tab:rank}), a constraint-violating solution might happen. In rare cases a rank-deficient $\mat{B}$ might produce a correct $y_1$ (e.~g. for $m=8, \sigma_{\nu}=10^{-6}$), but this is more the exception than the rule.
	\label{tab:stdy1}}
\begin{center}
	\begin{tabular}{|r|r||r|r|r|r|r|}
	\phline	
	 &$\sigma_{\nu}$&   	0& $10^{-10}$&  $10^{-8}$&  $10^{-6}$& $10^{-4}$ \\ \phline
$m$& & \multicolumn{5}{c|}{$\expec{y_1^2}$} \\ \phline\hline
 2&								&			1&			1&					1&					1&							1\\ \hline
 4&								&			1&			1&					1&					1&							1\\ \hline
 8&								&	3e-07&	4e-05&			3e-05&					1&							1\\ \hline
10&								&	1e-07&	3e-07&			2e-03&			5e-06&							1\\ \hline
12&								&	8e-08&	9e-05&			8e-04&					1&							1\\ \hline
20&								&	6e-08&	6e-05&			1e-04&			5e-01&							1\\ \hline
30&								&	4e-09&  1e-07&			6e-04&					1&							1\\ \hline
	\end{tabular}	
	%%%%
	%%%% generated as array 'stdy1' from sfa-tk/demo/arxiv2009_SFA2_rank.m 
	%%%% (using tau=1;nuf=25;q=1.2 in arxiv2009_SFA2.m)
	%%%%
\end{center}
}
\end{table}

\paragraph{Noise injection}

Another way of dealing with a singular or rank-deficient $\mat{B}$ is to add to the original time series a certain amount of noise, i.e. to perform a noise injection. If we replace for example
\begin{equation}
	w(t) \leftarrow w(t) + \nu(t)
	\label{eq:wNoise}
\end{equation}
where $\nu(t)$ is mean-free, normal-distributed noise with standard deviation $\sigma_{\nu}=10^{-4}$, then matrix $\mat{B}$ has full rank for all embedding dimensions $m=4,\ldots, 30$, thus GEN\_EIG will work as well as SVD\_SFA. (Of course the data are disturbed to a certain extent.) If we try to lower $\sigma_{\nu}$ to $10^{-6}, \ldots, 10^{-10}$ then $\mat{B}$ becomes gradually more rank-deficient (from 1\% to 50\% of the dimension of $\mat{B}$) and in parallel the performance of GEN\_EIG degrades in a roughly proportional way (see Tab.~\ref{tab:rank} and Tab.~\ref{tab:stdy1}). There might be certain cases where GEN\_EIG detects the correct slow signal with the correct constraints, but this can not be guaranteed.

\paragraph{Dependency on $\epsilon$} 

Is the cutoff parameter $\epsilon$ in \eqref{eq:closeToZero} for the 'close-to-zero'-condition of eigenvalues critical for the SVD\_SFA algorithm? One might suspect this to be the case since SFA in its slowest signal relies on the smallest eigenvalue. But first of all, the eigenvalues in \eqref{eq:closeToZero} refer to $\mat{B}$ while the slowest SFA signal has a small eigenvalue in $\mat{C'}=\expec{\dot{\vec{v}}\dot{\vec{v}}^T}$. Secondly, a small eigenvalue occuring in natural, noisy data will usually be not smaller than $10^{-3}$ while the 'close-to-zero'-condition is typically in the order of $10^{-7}\ldots 10^{-15}$ (machine accuracy) .  Nevertheles we tested  several other values $\epsilon=10^{-6},10^{-9},10^{-12}$ instead of the usual $10^{-7}$ and found virtually the same results. (Only $\epsilon$ as large as $10^{-5}$ resulted in a phase shift of the output signal.) Thus we conclude that the dependency on $\epsilon$ is {\em not} critical over a large range, at least not in our experiments.

\section{Conclusion}
\label{sec:Conclusion}

We have shown that the standard SFA algorithm GEN\_EIG based on generalized eigenvalues and implemented in the \textsc{Matlab} version of {\tt sfa-tk} yields under certain circumstances wrong results (in terms of the constraints and in terms of the slowness) for the slow SFA signals due to numerical instabilities. Those circumstances can be characterized as: "The covariance matrix  $\mat{B}$ of the expanded data is rank-deficient".

We have presented with SVD\_SFA a new algorithmic implementation which follows more closely the SFA approach of Wiskott and Sejnowski~\cite{WisSej2002} and which is numerically stable for all tested rank-deficient matrices $\mat{B}$. The \textsc{Matlab}-implementation is freely available for download.\footnote{see Appendix~B in Sec.~\ref{sec:AppendixB} for information how to download and use the extended package {\tt sfa-tk.V2}} With a certain trick we can avoid the direct representation of the {\em sphered} expanded data and thus can implement the new SVD\_SFA algorithm as efficiently as the original GEN\_EIG algorithm.   The new algorithm has roughly the same execution times as the old one since the time-consuming parts (expanding the data and accumulating the relevant matrices) remain the same. In our experiments the rank deficiency span a range between 1\% and  90\% of the number of dimensions in expanded space (see Tab.~\ref{tab:rank}). The single new parameter added by SVD\_SFA, namely the eigenvalue cutoff threshold $\epsilon$, was shown to be uncritical: the results obtained were insensitive to $\epsilon$-variation over a span of five decades. Thus the new algorithm does not need more parameter tuning than the old one, but it is more robust.

We have shown analytically that both implementations are equivalent as long as matrix  $\mat{B}$ is regular (has full rank).

An alternative approach to reach numerical stability is to avoid rank-deficient matrices by adding a certain amount of noise to the original data (noise injection). It has been shown that the original GEN\_EIG algorithm can be stabilized with the right amount of noise for all embedding dimensions $m$. Noise injection has however the drawback that certain parameters of the noise (noise amplitude, noise distribution and so on) have to be carefully tuned anew for each new SFA application. 

In this work the new algorithm SVD\_SFA has been applied only to driving force experiments. We plan to apply it also to SFA classification applications in the near future, where singular matrices might also occur in the case of smaller number of patterns per class. 

Although we have with SVD\_SFA now a robust algorithm, a general advice from the results presented here is to take always a look at rank$(\mat{B})$, the rank of the expanded data's covariance matrix, when performing SFA. Sometimes it might be worth to think about the possible reasons for a rank deficiency (e.g. too large $m$ or too few data) and to modify the experiment conditions accordingly.

\section{Appendix A}
\label{sec:AppendixA}
We review in this appendix some basic properties regarding covariance matrices and sphering matrices. 

Given is a set of $K$ data points $\vec{v} = [v_1,...,v_M]^T \in \Re^M$ with mean $\vec{v}_0 = \expec{\vec{v}}$. Here $\expec{\cdot}$ denotes the average over all $K$ points in the set. The covariance matrix associated with $\vec{v}$ is defined as
\begin{equation}
	\mat{B} = \Cov{\vec{v}} = \expec{(\vec{v}-\vec{v}_0)(\vec{v}-\vec{v}_0)^T} 
				  = \expec{\vec{v}\vec{v}^T}-\vec{v}_0\vec{v}_0^T
	\label{eq:covmat}
\end{equation}
Usually the covariance matrix deviates from the unit matrix because it has
\begin{enumerate}
	\item non-zero off-diagonal elements signaling dependencies between the dimensions of $\vec{v}$ and
	\item varying diagonal elements showing that different dimensions of $\vec{v}$ carry different amounts of the total variance of the signal. 
\end{enumerate}

\paragraph{Case 1: $\mat{B}=\Cov{\vec{v}}$ has no zero eigenvalues.} 

Sphering or whitening a set of data points $\vec{v}$ means to search a linear transformation $\mat{S}$ to obtain
\begin{equation}
	\vec{z} = \mat{S}(\vec{v}-\vec{v}_0) \quad\mbox{such that}\quad 
	\expec{\vec{z}}=0			\quad\mbox{and}\quad		\Cov{\vec{z}}=\expec{\vec{z}\vec{z}^T}=\mat{1}
	\label{eq:sphering}
\end{equation}
If no eigenvalues are zero then the sphering transformation $\mat{S}$ exists and is given by 
\begin{equation}
	\mat{S} = \mat{D}^{-1/2} \mat{R} = 
	\left(\begin{array}[]{ccc}
					1/\sqrt{\lambda_1} & & \\
					       & \cdots &      \\
					& & 1/\sqrt{\lambda_M}  
				\end{array}
	\right)	
	\left(\begin{array}[]{ccc}
					\cdots & \vec{r}_1^T& \cdots 	\\[1pt]
					       & \cdots &      		 		\\[1pt]
					\cdots & \vec{r}_M^T& \cdots 
				\end{array}
	\right)	= 
	\left(\begin{array}[]{ccc}
					\cdots & \vec{r}_1^T/\sqrt{\lambda_1}& \cdots \\[1pt]
					       & \cdots &      		 										\\[1pt]
					\cdots & \vec{r}_M^T/\sqrt{\lambda_M}& \cdots 
				\end{array}
	\right)	 
	\label{eq:sphere}
\end{equation}
where $\vec{r}_k$ is the $k$th eigenvector of $\Cov{\vec{v}}$ with eigenvalue $\lambda_k$ and unit length and where $\mat{R}$ is the matrix containing these eigenvectors in rows. $\mat{D}$ is the diagonal matrix of all eigenvalues. Usually, $\mat{D}$ and $\mat{R}$ are obtained by singular value decomposition (SVD).
One can easily verify that 
\begin{equation}
	\mat{R}\, \Cov{\vec{v}} \,\mat{R}^T = \mat{D}
	\label{eq:R}
\end{equation}
and with this
\begin{eqnarray}
	\expec{\vec{z}\vec{z}^T} &=& \mat{D}^{-1/2}
	 \mat{R}\, \expec{(\vec{v}-\vec{v}_0)(\vec{v}-\vec{v}_0)^T} \,\mat{R}^T \mat{D}^{-1/2} \nonumber\\
	&=& 
	\mat{D}^{-1/2} \underbrace{\mat{R}\, \Cov{\vec{v}} \,\mat{R}^T}_{=\mat{D}} \mat{D}^{-1/2} \nonumber\\
	&=& \mat{1}
	\label{eq:Ezz}
\end{eqnarray}
An obviously equivalent way of writing this is 
\begin{equation}
	\mat{S}\, \Cov{\vec{v}} \,\mat{S}^T = \mat{S}\, \mat{B} \,\mat{S}^T = \mat{1}
	\label{eq:S}
\end{equation}

\paragraph{Case 2: $\mat{B}=\Cov{\vec{v}}$ has eigenvalues $\lambda_j=0$.} 

The above sphering approach of course runs into problems if there are eigenvalues $\lambda_j=0$ or close to zero. The corresponding entries in $\mat{D}^{-1/2}$ would become infinity or very large and would amplify any noise or roundoff-errors multiplied with these entries. The standard trick from SVD~\cite{Press92} to deal with singular matrices $\mat{B}$ or $\mat{D}$ is to replace in $\mat{D}^{-1/2}$ each $1/\sqrt\lambda_j \approx 1/0$ with $0$ (!), thus effectively removing the corresponding eigenvector directions from further analysis. The condition for 'close to zero' is defined in relation to the largest eigenvalue
\begin{equation}
	\lambda_j/\lambda_{max} \leq \epsilon
	\label{eq:closeToZero}
\end{equation}
where we usually set $\epsilon=10^{-7}$. 
The matrix $\mat{S}$ becomes
\begin{equation}
	\mat{S} = \mat{D}^{-1/2} \mat{R} = 
	\left(\begin{array}[]{ccc}
					1/\sqrt{\lambda_1} & & \\
					       & \cdots &      \\
					& & 0  
				\end{array}
	\right)	
	\left(\begin{array}[]{ccc}
					\cdots & \vec{r}_1^T& \cdots 	\\[1pt]
					       & \cdots &      		 		\\[1pt]
					\cdots & \vec{r}_M^T& \cdots 
				\end{array}
	\right)	= 
	\left(\begin{array}[]{ccc}
					\cdots & \vec{r}_1^T/\sqrt{\lambda_1}& \cdots \\[1pt]
					       & \cdots &      		 										\\[1pt]
					\cdots & 0 & \cdots 
				\end{array}
	\right)	 
	\label{eq:sphere0}
\end{equation}
i.e. it has a $0$-row for each eigenvalue close to zero. $\mat{S}$ is not invertible, it projects into a subspace of dimension $P<M$ where $P$ is the number of non-zero eigenvalues of $\mat{B}$ (non-zero rows of $\mat{S}$).
The covariance matrix $\expec{\vec{z}\vec{z}^T}= \mat{S}\, \mat{B} \,\mat{S}^T $ in \eqref{eq:Ezz} becomes now a diagonal matrix with a $1$  for each non-zero eigenvalue and a $0$ for each eigenvalue close to zero. 

Our investigation above has shown that the results achieved with such a sphering approach are numerically stable in contrast to the generalized eigenvalue approach~\cite{Berkes2003} which has numerical problems with very small eigenvalues. 

\section{Appendix B: sfa-tk.V2}   %{\tt sfa-tk}}%.V2}}
\label{sec:AppendixB}
We briefly summarize in this appendix some information on how to obtain and use the extended package {\tt sfa-tk.V2}.
 
The package {\tt sfa-tk.V2} with the new SVD\_SFA algorithm can be downloaded from
\url{http://www.gm.fh-koeln.de/~konen/research/projects/SFA/sfa-tk.V2.zip}. It is a slightly modified version of Pietro Berkes' {\tt sfa-tk} which is available from  \url{http://people.brandeis.edu/~berkes/software/sfa-tk}~\cite{Berkes2003}. Installation of {\tt sfa-tk.V2} is the same as with {\tt sfa-tk} and all features of {\tt sfa-tk} are maintained.

{\tt sfa-tk.V2} contains -- besides the new SVD\_SFA algorithm -- also routines for SFA-classification, based on the ideas of~\cite{Berkes2005} together with Gaussian classifier routines.
Two new demo scripts, namely {\tt drive1.m} and {\tt class\_demo2.m} illustrate the usage of the new functionalities. A call of the new algorithm looks for example like
\begin{center}
\begin{minipage}{12cm}
\begin{verbatim}  
		 [y, hdl] = sfa2(x,'SVD_SFA');
\end{verbatim}
\end{minipage}
\end{center}

\vspace{1em}
The main modifications of {\tt sfa-tk.V2} are
\begin{itemize}
	\item {\tt lcov\_pca2.m}: covariance and sphering matrices are calculated with the more robust SVD method, same interface as {\tt lcov\_pca.m}.
	\item {\tt sfa\_step.m}: new parameter {\tt method}, handed over to {\tt sfa2\_step.m}.
	\item {\tt sfa2\_step.m}: 
	\begin{itemize}
		\item In step {\tt 'sfa'}: {\tt method='GEN\_EIG'} is the original \cite{Berkes2003}-code. {\tt method='SVD\_SFA'} is the new method along the lines of [WisSej02],  using {\tt lcov\_pca2.m}. If {\tt opts.dbg>0}, then both branches are executed and their results are compared with {\tt sfa2\_dbg.m}.  
		\item In step {\tt 'expansion'}: {\tt method='TIMESERIES'} is the original \cite{Berkes2003}-code where the goal is to minimize the time difference signal. The {\tt method='CLASSIF'} is new for classification purposes, along the lines of~\cite{Berkes2005}. Each data chunk is assumed to be a set of patterns from the same class, and the goal is to minimize the pairwise pattern difference.
	\end{itemize} 
	\item {\tt sfa2.m}: new parameters {\tt method} and {\tt pp\_type}.
	\item {\tt sfa2\_dbg.m}: perform certain debug checks and print out diagnostic information.
	\item {\tt drive1.m}: demo script for performing the driving force demo experiments.
	\item {\tt class\_demo2.m}: demo script for performing classification experiments along the lines of~\cite{Berkes2005}. Instead of handwritten digits we use the Vowel benchmark dataset from UCI
machine learning repository (\url{http://archive.ics.uci.edu/ml})
\end{itemize} 
The full list of modifications is available in file {\tt CHANGES.htm} in the download package {\tt sfa-tk.V2.zip}.

\section{Acknowledgment}

 I am grateful to Laurenz Wiskott for helpful discussions on SFA and to Pietro Berkes for
 providing the \textsc{Matlab} code for {\tt sfa-tk}~\cite{Berkes2003}.  This work
 has been supported by the Bundesministerium f\"ur Forschung und Bildung (BMBF) under the grant 
 SOMA (AIF FKZ 17N1009, "Systematic Optimization of Models in IT and Automation") and by the Cologne University of Applied Sciences under the research focus grant COSA.

% --- this is the normal way: run BibTex on SFA.bib to produce main.bbl:
%\bibliographystyle{alpha}
%\bibliography{../arxiv2009.SFA/SFA}

% --- bititems.tex is a copy of main.bbl, which is needed for arXiv.org (which does not run BibTex)

\end{document}